\documentclass{article}

\PassOptionsToPackage{numbers, compress}{natbib}

\usepackage[dblblindworkshop, final]{neurips_2026}
\workshoptitle{Foundation Models for the Brain and Body}

\usepackage[utf8]{inputenc} 
\usepackage[T1]{fontenc}    
\usepackage{hyperref}       
\usepackage{url}            
\usepackage{booktabs}       
\usepackage{amsfonts}       
\usepackage{nicefrac}       
\usepackage{microtype}      
\usepackage{xcolor}         
\usepackage{graphicx}
\usepackage{amsmath}
\usepackage{etoolbox}
\usepackage{makecell}
\usepackage{tabularx}
\usepackage{multirow}
\usepackage{bm}
\usepackage{wrapfig}
\newcommand{\comment}[1]{}

\newcommand{\ourmethod}{BREM-NET}
\newcommand{\uourmethod}{U-\ourmethod}
\newcommand{\lourmethod}{L-BREM-NET}

\title{Learning Behaviorally Relevant Dynamics in Multimodal Neural Recordings}

\author{%
  Parastoo Azizeddin$^{1}$ \quad
  Eray Erturk$^{1}$ \quad
  Maryam M. Shanechi$^{*1\text{--}4}$ \\
  \normalfont University of Southern California, Los Angeles, CA \\
  \normalfont\texttt{\{azizeddi,eerturk,shanechi\}@usc.edu}
}

\begin{document}

\maketitle
\begingroup
\renewcommand{\thefootnote}{$*$}\footnotetext{Corresponding author: \texttt{shanechi@usc.edu}.}
\renewcommand{\thefootnote}{1}\footnotetext{Ming Hsieh Department of Electrical and Computer Engineering, University of Southern California}
\renewcommand{\thefootnote}{2}\footnotetext{Thomas Lord Department of Computer Science, University of Southern California}
\renewcommand{\thefootnote}{3}\footnotetext{Alfred E. Mann Department of Biomedical Engineering, University of Southern California}
\renewcommand{\thefootnote}{4}\footnotetext{Neuroscience Graduate Program, University of Southern California}
\endgroup

\begin{abstract}
Multimodal neural data can enable a more complete understanding of brain dynamics underlying behavior. Modalities such as spiking activity, local field potentials (LFPs), and behavior capture diverse spatiotemporal aspects of brain processes, yet prior neural-behavioral models that separate sources of variability rely on a single neural modality. By leveraging these complementary strengths, multimodal neural fusion can provide a unified, rich representation of brain-behavior processes and address the limitations of single-modality analyses.
Here we develop \ourmethod{}, a nonlinear multimodal dynamical model that integrates behavior with multiple neural modalities having distinct statistical characteristics and temporal resolutions. \ourmethod{} performs multimodal neural fusion during inference while separating behaviorally relevant and residual neural dynamics via a structured learning framework. In a public multimodal non-human primate dataset with simultaneous spiking and local field potential recordings during a 2D reaching task, \ourmethod{} separates these two types of dynamics and learns them more accurately, as reflected in better behavior decoding, better neural prediction, and stronger latent alignment with behavior, and it retains these capabilities when neural modalities are asynchronous or suffer dropouts, which are major challenges in real-world recordings. This framework provides a new tool for studying behaviorally relevant neural computations across spatiotemporal scales of brain activity.
\end{abstract}

\section{Introduction}
Understanding neural dynamics across spatiotemporal scales and their relationship to behavior is a fundamental challenge in neuroscience, and different scales are captured by different modalities: spikes reflect rapid transient events of individual neurons and are often modeled as Poisson, LFPs capture slower oscillatory processes at a larger spatial scale and are typically modeled as Gaussian, and behavior is a high-level output of the nervous system with complex relationships to neural activity~\citep{pesaran2002temporal,einevoll2013modelling,buzsaki2012origin,hsieh2018multiscale,gallego2020long,ramezani2024innovating,sharafi2026population,householder2026extraocular,torabi2026neuromamballm}. Integrating them can provide a unified dynamical framework for uncovering the multimodal neural mechanisms driving behavior and improve neural-behavioral prediction, e.g., for brain-computer interfaces (BCIs), where spike signals often degrade over time while LFPs remain comparatively stable~\citep{stavisky2015high}. However, developing such models remains challenging. First, neural modalities have both \emph{behaviorally relevant} dynamics shared with behavior and other neural dynamics that these leave unexplained, which we refer to as \emph{residual neural dynamics}; separating them enables new scientific analyses -- such as characterizing which aspects of multimodal neural activity are sufficient to describe behavior, apart from the remaining neural variability -- and requires functionally separated latent subspaces rather than a single unified representation. Second, discrete spikes and continuous LFPs have distinct statistical characteristics. Third, in real recordings modalities are often asynchronous, i.e., sampled at different temporal resolutions, or suffer dropouts. Finally, their dynamics can be nonlinear.

Prior work falls into two categories, neither of which addresses all of the above challenges. The first jointly models a \emph{single} neural modality with behavior to learn behaviorally relevant dynamics, using linear and nonlinear models~\citep{kobak2016demixed,sani2021modeling,vahidi2024modeling,hurwitz2021targeted, sani2024dissociative,schneider2023learnable,gondur2023multi,zhou2020learning,whiteway2021partitioning,wang2024exploring,wu2025disentangling,hosseini2025dynamical}; these separate the two types of dynamics but do not fuse multiple neural modalities in modeling or inference. The second fuses multiple neural time-series~\citep{coleman2011mixed,zhou2020learning,abbaspourazad2021multiscale,singh2021neural,kramer2021reconstructing,rezaei2023inferring,ahmadipour2024multimodal,kim2026unsupervised,erturk2025dynamical} -- but does not separate behaviorally relevant from residual neural dynamics, so the two can be conflated. Further, many of these methods are designed for fusing modalities with the same temporal resolution.

We propose \textbf{B}ehaviorally \textbf{Re}levant modeling of \textbf{M}ultimodal \textbf{Ne}ural \textbf{T}ime-series (\textbf{\ourmethod{}}), a nonlinear multimodal dynamical model that simultaneously: 1) incorporates multiple neural modalities and behavior in a unified model; 2) learns behaviorally relevant neural dynamics from multimodal neural time-series first and then models the residual neural dynamics in a separate set of latents via a structured learning framework; and 3) supports fusion of modalities with heterogeneous distributions, different temporal resolutions, or dropouts. On a public multimodal non-human primate (NHP) dataset with simultaneous spiking and LFP recordings during a 2D reaching task~\citep{o_doherty_2020_3854034}, and in simulations, \ourmethod{} fuses information across neural modalities -- even when sampled asynchronously -- while separating the two types of dynamics. Doing so, it learns more accurate behaviorally relevant latents, as reflected in behavior decoding and alignment with behavior, and achieves accurate neural prediction via its residual neural latents, which capture the neural variability that the behaviorally relevant latents leave unexplained. This provides a new tool for investigating multiscale neural computations underlying behavior, with implications for multimodal BCIs under realistic, asynchronous conditions.

\section{Methods}\label{methods section}
\begin{figure*}[!t]
  \centering
  \includegraphics[width=\textwidth]{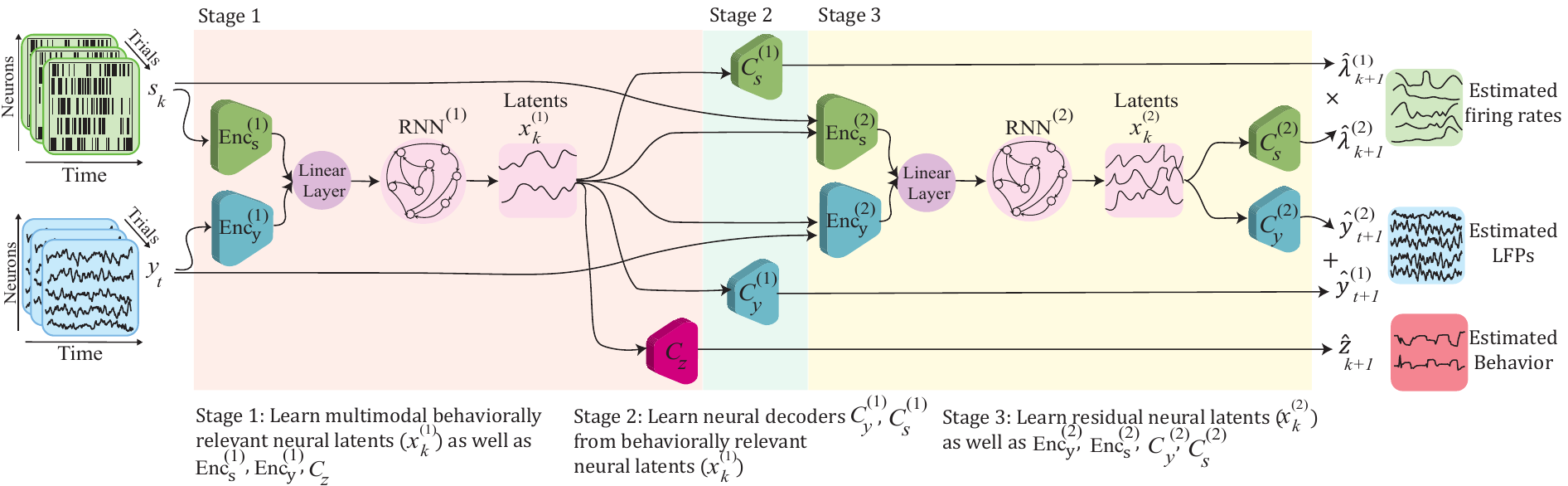}
  \caption{\textbf{\ourmethod{} model architecture.} At each time step, multimodal neural observations are used to infer latent states that predict both behavior and neural activity. \textbf{Stage 1} learns behaviorally relevant latent states $\mathbf{x}_k^{(1)}$ from multimodal neural data by jointly training an RNN with modality-specific encoders and a behavior decoder to minimize the negative log-likelihood (NLL) of behavior. \textbf{Stage 2} fits two modality-specific neural decoders to minimize the NLL of the neural time-series given $\mathbf{x}_k^{(1)}$. \textbf{Stage 3} learns residual neural latent states $\mathbf{x}_k^{(2)}$ by jointly fitting a second RNN with modality-specific encoders and decoders, minimizing the multimodal NLL of neural observations given both latent sets, so that it learns residual neural dynamics not predicted by $\mathbf{x}_k^{(1)}$.}
  \label{architecture}
\end{figure*}
The architecture of \ourmethod{} (Fig.~\ref{architecture}) is designed for joint modeling of behavior and multiple neural modalities observed simultaneously, and addresses two key challenges: separating the behaviorally relevant from the residual neural dynamics, and handling asynchronous modalities. For the separation, \ourmethod{} employs two RNNs trained through a structured three-stage process: the first RNN learns behaviorally relevant neural latents in Stage 1, whose mapping to the neural modalities is learned in Stage 2, while the second RNN learns residual neural latents in Stage 3 that capture neural dynamics unexplained by the first. To handle asynchronous modalities, we introduce separate modality-specific encoders that process each modality at its native temporal resolution.

\subsection{Model formulation}
We model the multimodal neural and behavior time-series as observations of a nonlinear
latent dynamical system. The system state $\mathbf{x}_k \in \mathbb{R}^{n_x}$ evolves in
time through a nonlinear recursion with additive noise, $\mathbf{x}_{k+1} =
F(\mathbf{x}_{k}) + \mathbf{w}_k$, and each observed time-series is generated from the
state at the corresponding time step, conditionally independently of the others. Concretely,
the state is passed through a modality-specific nonlinear decoder network,
$\bm{\bar{C}}_\mathbf{y}(\cdot)$, $\bm{\bar{C}}_\mathbf{s}(\cdot)$ or
$\bm{\bar{C}}_\mathbf{z}(\cdot)$, whose output parameterizes a modality-specific likelihood:
$\mathbf{y}_t \sim p_\mathbf{y}(\cdot \mid \bm{\bar{C}}_\mathbf{y}(\mathbf{x}_t))$ for the
LFP, $\mathbf{s}_k \sim p_\mathbf{s}(\cdot \mid \bm{\bar{C}}_\mathbf{s}(\mathbf{x}_k))$ for
the spiking activity, and $\mathbf{z}_k \sim p_\mathbf{z}(\cdot \mid
\bm{\bar{C}}_\mathbf{z}(\mathbf{x}_k))$ for behavior (Appendix~\ref{app:Method Formulation}).
Because each modality keeps its own likelihood, sources with heterogeneous statistics are
modeled jointly: we use Poisson for spike counts and Gaussian for LFP and for the continuous
behavior signals, and the same formulation accommodates other choices such as Bernoulli for
binary outcomes.

Here $\mathbf{s}_k$ is indexed by $k$ 
and $\mathbf{y}_t$ by $t$. The distinct
index sets emphasize that the two neural modalities can be sampled at different rates, with
$k$ indexing the faster one; we perform latent state inference on that faster timescale,
which enables more flexible behavior decoding. The behavior time-series $\mathbf{z}_k \in
\mathbb{R}^{n_z}$ is used only during learning and is decoded from the latent states at
inference time.

We jointly and nonlinearly fuse the two neural time-series to infer the latent states using an RNN, $\mathbf{x}_{k+1} = A( \mathbf{x}_{k}) + \mathbf{Enc_y}(\mathbf{y}_t) + \mathbf{Enc_s}(\mathbf{s}_k)$,
where $A(\cdot)$ captures the temporal evolution of the latents and $\mathbf{Enc_y}(\cdot)$, $\mathbf{Enc_s}(\cdot)$ are modality-specific nonlinear encoders. This lets \ourmethod{} leverage complementary information across the neural modalities in a shared latent space, while the separate modality-specific encoders address the different temporal resolutions and modality-specific dropouts common in real-world recordings: if one modality is unavailable at a given $k$, the corresponding encoder output is set to zero, so the state update relies only on the available modality (Appendix~\ref{app:Method Details}).

Beyond multimodal fusion, another major goal of \ourmethod{} is to functionally separate the behaviorally relevant and the residual neural dynamics within the multimodal neural time-series. Following~\citep{sani2021modeling,hurwitz2021targeted,sani2024dissociative}, we define two sets of latent states: behaviorally relevant neural latents $\mathbf{x}_k^{(1)} \in \mathbb{R}^{n_1}$, which capture the dynamics shared between behavior and the multimodal neural time-series, and residual neural latents $\mathbf{x}_k^{(2)} \in \mathbb{R}^{n_x-n_1}$, which encode the residual neural dynamics. We then expand the formulation into the following structured form:
\begin{equation}
\begin{aligned}
\mathbf{x}_{k+1}^{(1)} &= f^{(1)}\!\big(\mathbf{x}_k^{(1)}, \mathbf{y}_t, \mathbf{s}_k\big) = A^{(1)}(\mathbf{x}_k^{(1)}) + \mathbf{Enc}^{(1)}_{\mathbf{y}}(\mathbf{y}_t) + \mathbf{Enc}^{(1)}_{\mathbf{s}}(\mathbf{s}_k), \\
\mathbf{x}_{k+1}^{(2)} &= f^{(2)}\!\big(\mathbf{x}_k^{(2)}, \mathbf{y}_t, \mathbf{s}_k, \mathbf{x}_{k+1}^{(1)}\big) = A^{(2)}(\mathbf{x}_k^{(2)}) + \mathbf{Enc}^{(2)}_{\mathbf{y}}(\mathbf{y}_t,\mathbf{x}_{k+1}^{(1)}) + \mathbf{Enc}^{(2)}_{\mathbf{s}}(\mathbf{s}_k,\mathbf{x}_{k+1}^{(1)}), \\
\mathbf{\hat{y}}_t &= \bm{C}_{\mathbf{y}}^{(1)}(\mathbf{x}_t^{(1)}) + \bm{C}_{\mathbf{y}}^{(2)}(\mathbf{x}_t^{(2)}), \qquad
\mathbf{\hat{\lambda}}_k = \bm{C}_{\mathbf{s}}^{(1)}(\mathbf{x}_k^{(1)}) \times \bm{C}_{\mathbf{s}}^{(2)}(\mathbf{x}_k^{(2)}), \qquad
\mathbf{\hat{z}}_k = \bm{C}_{\mathbf{z}}(\mathbf{x}_k^{(1)})
\end{aligned}
\label{eq:equation_formulation_expanded}
\end{equation}
where $f^{(1)}(\cdot)$ and $f^{(2)}(\cdot)$ denote the RNNs describing the behaviorally relevant and the residual neural dynamics, each parameterized by a recursion network $A(\cdot)$ and modality-specific encoders with the corresponding superscript, and $\bm{C}_\mathbf{y}(\cdot)$, $\bm{C}_\mathbf{s}(\cdot)$, $\bm{C}_\mathbf{z}(\cdot)$ are modality-specific decoders: $\mathbf{\hat{\lambda}}_k$ is the predicted Poisson firing rate, $\mathbf{\hat{y}}_t$ the predicted Gaussian modality, and $\mathbf{\hat{z}}_k$ the predicted behavior. Behavior is predicted from the behaviorally relevant latents alone, whereas the neural modalities are predicted from both latent sets, whose contributions are fused by adding the Gaussian components and multiplying the Poisson firing rates, equivalently summing log-rates (Appendix~\ref{app:Method Formulation}). Since $\mathbf{x}_{k+1}$ is estimated purely from past neural data, the model supports causal prediction, and passing $\mathbf{x}_{k+1}^{(1)}$ to the second RNN makes it learn the residual dynamics not already captured by the first.

\subsection{Structured learning of behaviorally relevant and residual neural dynamics}
To functionally separate the two types of dynamics, we learn the model in Eq.~\ref{eq:equation_formulation_expanded} with a structured three-stage procedure, in which each stage optimizes a subset of the parameters while holding those learned earlier fixed (Appendix~\ref{app:Learning Parameters}).

\textbf{Stage 1.} We train the RNN $f^{(1)}(\cdot)$ jointly with the behavior decoder $\bm{C}_\mathbf{z}(\cdot)$, minimizing the negative log-likelihood (NLL) of behavior given the inferred latents $\mathbf{x}_k^{(1)}$:
\begin{equation}
\mathbf{L_{z}} = \sum_{k} \mathbf{NLL}\big( \mathbf{z}_k; \bm{C}_\mathbf{z}(\mathbf{x}^{(1)}_k) \big).
\label{eq:loss_stage1}
\end{equation}
This ensures these latents capture the behaviorally relevant information in the multimodal neural data.

\textbf{Stage 2.} We train the neural decoders $\bm{C}_\mathbf{y}^{(1)}(\cdot)$ and $\bm{C}_\mathbf{s}^{(1)}(\cdot)$, which predict the neural modalities from $\mathbf{x}_k^{(1)}$, by minimizing the combined NLL of both modalities under their respective likelihoods, with a scaling factor $\alpha$ that balances their scale differences (Appendix~\ref{alpha_details}):
\begin{equation}
\mathbf{L}^{(1)}_{\mathbf{y},\mathbf{s}} = \sum_{t} \mathbf{NLL_y}\big( \mathbf{y}_t;\bm{C}_\mathbf{y}^{(1)}(\mathbf{x}_t^{(1)}) \big) + \alpha \sum_{k} \mathbf{NLL_s}\big( \mathbf{s}_k; \bm{C}_\mathbf{s}^{(1)}(\mathbf{x}_k^{(1)})\big).
\label{eq:loss_stage2}
\end{equation}

\textbf{Stage 3.} Finally, we train the second RNN $f^{(2)}(\cdot)$ jointly with the decoders $\bm{C}_\mathbf{y}^{(2)}(\cdot)$ and $\bm{C}_\mathbf{s}^{(2)}(\cdot)$, minimizing the combined NLL of both neural modalities given \emph{both} $\mathbf{x}_k^{(1)}$ and $\mathbf{x}_k^{(2)}$. Because Stages 1--2 already account for the part of $\mathbf{y}_t$ and $\mathbf{s}_k$ predictable from the behaviorally relevant latents, this stage explains the residual neural dynamics:
\begin{equation}
\mathbf{L}^{(2)}_\mathbf{y,s} = \sum_{t} \mathbf{NLL_y}\big(\mathbf{y}_t; \bm{C}_\mathbf{y}^{(1)}(\mathbf{x}_t^{(1)}), \bm{C}_\mathbf{y}^{(2)}(\mathbf{x}_t^{(2)}) \big) + \alpha \sum_{k} \mathbf{NLL_s}\big(\mathbf{s}_k; \bm{C}_\mathbf{s}^{(1)}(\mathbf{x}^{(1)}_k), \bm{C}_\mathbf{s}^{(2)}(\mathbf{x}_k^{(2)})\big).
\label{eq:loss_stage3}
\end{equation}

\section{Experimental Results}
\begin{figure*}[!ht]
\begin{center}
\centerline{\includegraphics[width=0.70\textwidth]{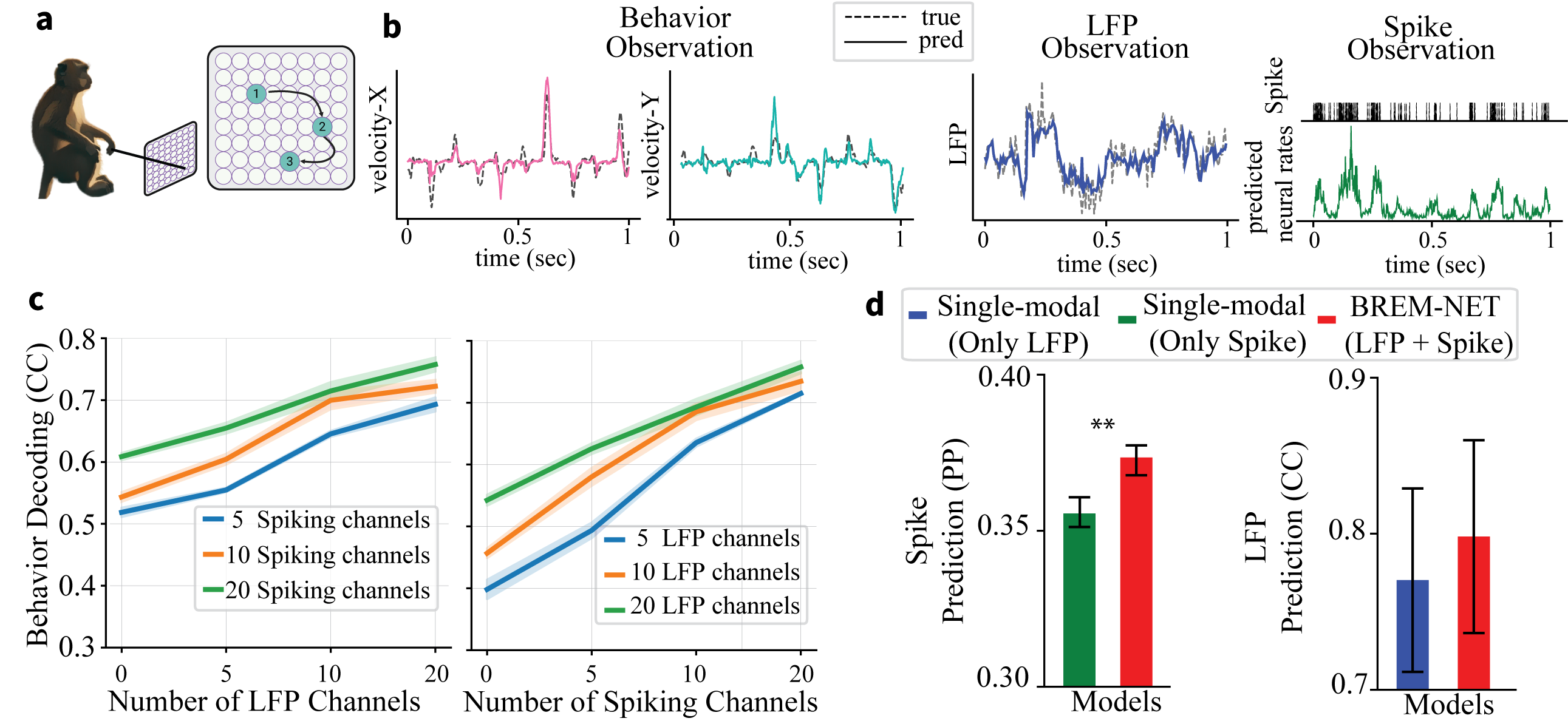}}
\caption{\textbf{\ourmethod{} successfully fuses multimodal neural time-series, thus improving neural-behavioral predictions in the NHP grid reaching dataset.} \textbf{(a)} Dataset and task. \textbf{(b)} True behavior, LFP and spike time-series (dashed) versus their \ourmethod{} predictions (solid) for a representative session. \textbf{(c)} Cross-validated behavior decoding CC when 5, 10 or 20 spiking (\emph{left}) or LFP (\emph{right}) channels were the primary modality and channels of the other modality were fused; lines show the mean, shaded areas the SEM. \textbf{(d)} Cross-validated spike prediction PP (\emph{left}) and LFP prediction CC (\emph{right}) with \ourmethod{} vs.\ single-modal models. N=20; **: $p<0.001$, one-sided Wilcoxon signed-rank test.}
\label{fig:sabes_results_all}
\end{center}
\end{figure*}
\textbf{Dataset and metrics.} We validated \ourmethod{} on a public NHP dataset~\citep{o_doherty_2020_3854034} with simultaneous spiking activity and raw LFP from primary motor cortex during sequential 2D reaches to random targets in virtual reality (Fig.~\ref{fig:sabes_results_all}a, Appendix~\ref{sabes_data}), and on simulated Gaussian and Poisson systems (Appendix~\ref{app:simulation_section}). All results use 5-fold cross-validation across all sessions. For Gaussian signals we report the Pearson correlation coefficient (CC); for spike counts, prediction power ($\mathrm{PP} = 2\,\mathrm{AUC}-1$)~\citep{macke2011empirical,truccolo2010collective}; both averaged across dimensions (Appendix~\ref{app:evaluation metric}).

\subsection{Multimodal fusion extracts more accurate behaviorally relevant neural latents}\label{sabes results_all}
We treated spike counts and raw LFPs as the neural time-series and 2D cursor velocity as behavior, with $n_x = 64$ and $n_1 = 16$ selected by sweeping both dimensionalities on a single session (Appendix~\ref{hyperparams}). To assess fusion, either spikes or LFPs served as the primary modality while increasing numbers of channels of the other were fused (Fig.~\ref{fig:sabes_results_all}c). Adding LFP channels to spiking activity consistently improved behavior decoding ($p<10^{-5}$, $n=20$, one-sided Wilcoxon signed-rank test), and a similar trend held when LFPs were primary, with larger gains in that case, suggesting spikes encoded more behavior-related information here. More generally, fusion yields its largest gains when the primary modality is information-limited (e.g., fewer primary channels in Fig.~\ref{fig:sabes_results_all}c), the regime characteristic of chronic BCIs in which spike quality degrades over time while LFPs remain comparatively stable~\citep{stavisky2015high}. Fusion also improved neural prediction of each modality over single-modal models (Fig.~\ref{fig:sabes_results_all}d).

\begin{figure}[!ht]
\begin{center}
\centerline{\includegraphics[width=0.64\textwidth]{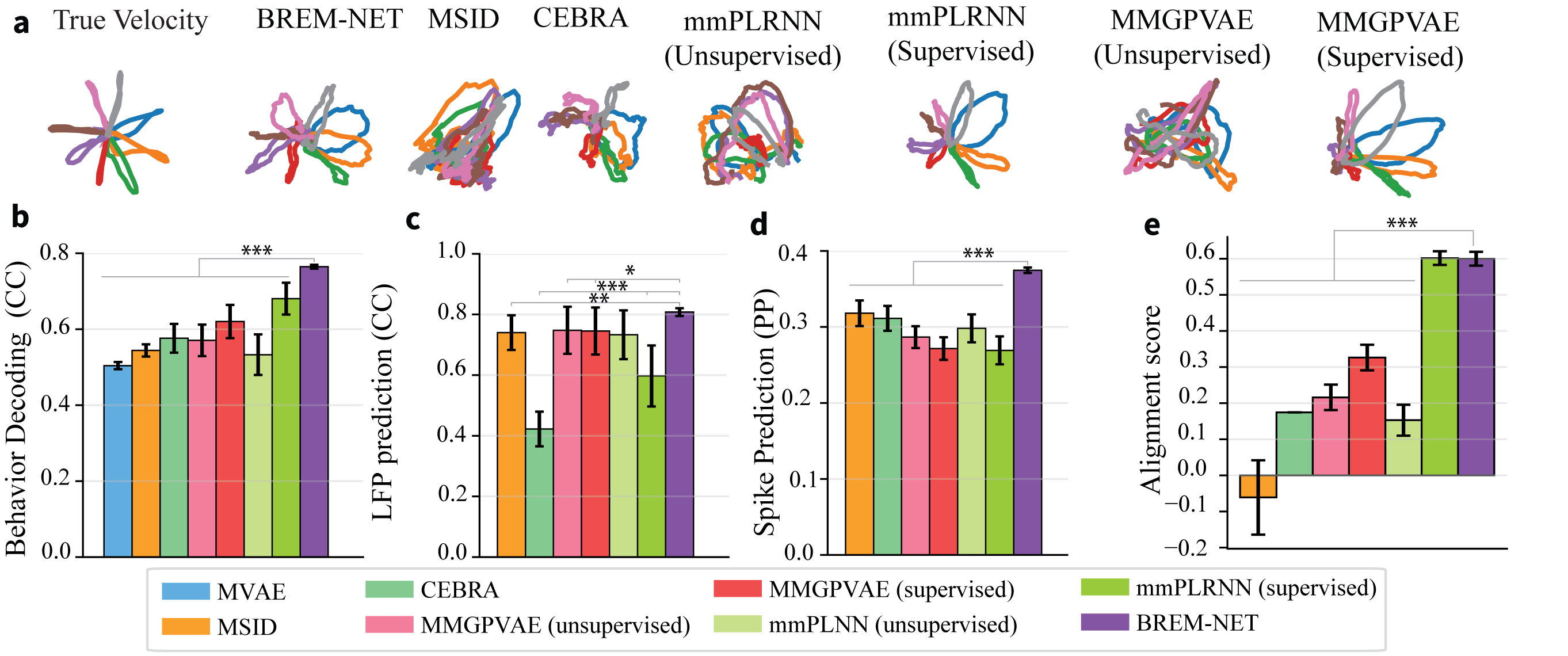}}
\caption{\textbf{Latents inferred by \ourmethod{} are more aligned with true behavior trajectories and achieve accurate neural-behavioral prediction.} \textbf{(a)} True velocity trajectories vs.\ condition-averaged latent trajectories for \ourmethod{} and baselines. \textbf{(b)} Behavior decoding CC. \textbf{(c)} LFP prediction CC. \textbf{(d)} Spike prediction PP. \textbf{(e)} Alignment score. Bars show mean $\pm$ SEM (N=20); *: $p<0.05$, **: $p<0.001$, ***: $p<10^{-7}$, one-sided Wilcoxon signed-rank test.}
\label{fig:nhp basleines barplot}
\end{center}
\end{figure}

To assess the functional separation of the two sets of latents, we evaluated their specialization by using $\mathbf{x}^{(1)}$ and $\mathbf{x}^{(2)}$ separately (Fig.~\ref{app:x1_x2_both_preds}b, Appendix~\ref{app:separation}): $\mathbf{x}^{(1)}$ accurately decoded behavior, while $\mathbf{x}^{(2)}$ excelled at predicting neural activity but decoded behavior substantially less accurately. As a complementary geometric characterization, true velocity trajectories were much more aligned with $\mathbf{x}^{(1)}$ than with $\mathbf{x}^{(2)}$ trajectories (Fig.~\ref{app:x1_x2_both_preds}c), as quantified by an alignment score, defined as the Pearson correlation between the Euclidean distance matrices of the behavior signals and the neural latent trajectories~\citep{jiang2025extracting}. Together, these results indicate that Stage 1 captures the behaviorally relevant dynamics while Stage 3 captures the residual neural variability.

\textbf{Comparison to multimodal baselines.} We compared \ourmethod{} to MVAE~\citep{wu2018multimodal}, MSID~\citep{ahmadipour2024multimodal}, multi-session CEBRA~\citep{schneider2023learnable}, mmPLRNN~\citep{kramer2021reconstructing}, and MMGPVAE~\citep{gondur2023multi} (Fig.~\ref{fig:nhp basleines barplot}; Appendix~\ref{baselines}). For fairness, mmPLRNN and MMGPVAE received the two neural modalities as inputs and were trained both unsupervised (their original form) and supervised, with an added behavior reconstruction loss. As shown in Fig.~\ref{fig:nhp basleines barplot}, \ourmethod{}'s behaviorally relevant latents were more aligned with true velocity than those of all baselines (Fig.~\ref{fig:nhp basleines barplot}a,e), and it decoded behavior and predicted neural activity more accurately (Fig.~\ref{fig:nhp basleines barplot}b--d).

\subsection{Multimodal fusion and structured learning are both critical}
We ablated \ourmethod{}'s components (Fig.~\ref{fig:ablation_figure}, Appendix~\ref{app:ablations}). Multimodal fusion, behavior supervision and the staged procedure each contribute: latents from the single-modal variants, from the unsupervised \uourmethod{} and from a single-stage joint-loss variant are all markedly less aligned with behavior, and \ourmethod{} also beats its linear counterpart \lourmethod{}. 

\subsection{Robustness to asynchronous sampling and modality dropouts}
We evaluated the robustness of \ourmethod{} when neural modalities are partially dropped out or sampled at different rates (different temporal resolutions). First, in the NHP grid-reaching dataset, we downsampled LFPs by a factor of 5. As shown in Tab.~\ref{tab:diff_timescales}, \ourmethod{} maintained its behavior decoding performance in this asynchronous setting (\(p>0.5\), \(n=20\), two-sided Wilcoxon signed-rank test), confirming that it does not require the modalities to be strictly aligned in time. Furthermore, under stochastic modality-specific dropouts ($q \in \{0.2,0.4,0.6,0.8\}$ per timestep), we show that behavior decoding declined by less than 2\% even with 80\% of LFP samples removed and by about 5\% at 40\% spike dropout (Appendix~\ref{different resolution}). Overall, these results show that \ourmethod{} can robustly fuse even asynchronous and incomplete neural modalities.

\section{Discussion}

We introduced \ourmethod{}, a nonlinear dynamical model that integrates multiple neural modalities alongside behavior, using modality-specific encoders and decoders and separate behaviorally relevant and residual neural latents. In a multimodal NHP dataset and in simulations, our method functionally separated behaviorally relevant and residual neural dynamics and learned them more accurately, leading to better neural and behavior inferences, even when modalities had distinct temporal resolutions or modality-specific dropouts, which are natural challenges in real-world applications such as BCIs. Our ablations showed that these benefits arise from multimodal modeling and supervised structured learning.

By first learning a compact latent state that is predictive of behavior from multiple neural modalities, and then modeling the remaining neural dynamics in a separate set of latents, \ourmethod{} provides a tool to study how neural activity at different spatiotemporal scales jointly relates to behavior. For example, it can quantify how each modality contributes to behaviorally relevant dynamics and how fusion changes this contribution, as in our analyses of spikes and LFPs (Fig.~\ref{fig:sabes_results_all}c), while its compact behaviorally relevant state supports accurate behavior decoding for BCIs.

\textbf{Limitations and future directions.} We applied our method to spikes and raw LFPs given their importance in neuroscience and BCIs. Important future directions include extending it to more modalities, such as EEG~\citep{azizeddin2026multimodal}, intracranial EEG~\citep{vaziri2025nonlinear}, or physiological signals such as heart rate, and to other brain regions and behaviors~\citep{jha2026prioritized}.

\begin{ack}
This work was supported by National Institutes of Health (NIH) grants R61MH135407, R01MH123770, and RF1DA056402, the One Mind Rising Star Award, and the Foundation for OCD Research (FFOR).
\end{ack}

\small
\bibliographystyle{unsrt}
\bibliography{example_paper}
\normalsize
\newpage
\appendix
\renewcommand{\theHequation}{A.\theequation}
\renewcommand{\theHtable}{A.\thetable}
\renewcommand{\theHfigure}{A.\thefigure}
\setcounter{table}{0}
\setcounter{figure}{0}
\setcounter{equation}{0}
\renewcommand{\thetable}{A.\arabic{table}}
\renewcommand{\thefigure}{A.\arabic{figure}}
\renewcommand{\theequation}{A.\arabic{equation}}

\section{Appendix}
\subsection{Method Details}\label{app:Method Details}
\subsubsection{Method Formulation} \label{app:Method Formulation}
In Equation~\ref{eq:equation_formulation_expanded}, we provided a compact formulation of our RNN-based architecture. Here, we present a more detailed formulation that explicitly describes the latent dynamics and the associated model components used in \ourmethod{}. The full model is defined as:

\begin{equation}
\left\{
\begin{aligned}
\begin{bmatrix}
\mathbf{x}_{k+1}^{(1)} \\
\mathbf{x}_{k+1}^{(2)}
\end{bmatrix}
&=
\begin{bmatrix}
A^{(1)}(\mathbf{x}_k^{(1)})  \\
A^{(2)}(\mathbf{x}_k^{(2)})
\end{bmatrix} 
+
\begin{bmatrix}
\mathbf{Enc}^{(1)}_\mathbf{y}(\mathbf{y}_t) + \mathbf{Enc}^{(1)}_\mathbf{s}(\mathbf{s}_k) \\
\mathbf{Enc}^{(2)}_\mathbf{y}(\mathbf{y}_t, \mathbf{x}_{t+1}^{(1)}) + \mathbf{Enc}^{(2)}_\mathbf{s}(\mathbf{s}_k, \mathbf{x}^{(1)}_{k+1})
\end{bmatrix} \\
\mathbf{\hat{y}}_t &= \bm{C}_\mathbf{y}^{(1)}(\mathbf{x}_t^{(1)}) +\bm{C}_\mathbf{y}^{(2)}(\mathbf{x}_t^{(2)}) \\
\mathbf{\hat{\lambda}}_k &= \bm{C}_\mathbf{s}^{(1)}(\mathbf{x}_k^{(1)}) \times \bm{C}_\mathbf{s}^{(2)}(\mathbf{x}_k^{(2)}) \\
\mathbf{\hat s}_k &\sim \mathcal{P}ois(\mathbf{\hat{\lambda}}_k) \\
\mathbf{\hat{z}}_k &= \bm{C_z}(\mathbf{x}_k^{(1)})
\label{eq:equation appendix}
\end{aligned}
\right.
\end{equation}
where \( A^{(1)}(\cdot)\), and \( A^{(2)}(\cdot)\) represents the state transition function, capturing the temporal evolution of the latent state, and \ \(\mathbf{Enc}^{(1)}_\mathbf{y}(\cdot)\),  \(\mathbf{Enc}^{(1)}_\mathbf{s}(\cdot)\),  \(\mathbf{Enc}^{(2)}_\mathbf{y}(\cdot)\), and \(\mathbf{Enc}^{(2)}_\mathbf{s}(\cdot)\) are modality-specific encoders for \(\mathbf{y}_t\) and \( \mathbf{s}_k\), respectively. We use distinct time indices \(t\) and \(k\) for \(y\) and \(s\) respectively to indicate that neural modalities can be sampled at different temporal resolutions with k indexing the faster sampling rate.
At each time step \(k\), the latent state is updated by combining the predicted state evolution with contributions from any available modalities through their corresponding encoders. When a modality is not observed at a given time step (e.g., due to asynchronous sampling or dropout), its encoder contribution is  set to zero, such that no update is introduced from that modality. Consequently, the latent state is inferred using the prior prediction together with any available modalities at that time step. This mechanism is analogous to Kalman filtering~\citep{brookner1998tracking} with missing observations, where the Kalman gain for unobserved measurement dimensions becomes zero (equivalently, assuming infinite observation noise), resulting in no correction from those dimensions and relying solely on the prediction and available measurements.

Here, the overall latent state \(\mathbf{x}_k \in \mathbb{R}^{n_x}\), which captures the underlying dynamics of the neural-behavioral data, is designed to separate behaviorally relevant neural dynamics, represented by \(\mathbf{x}^{(1)}_k \in \mathbb{R}^{n_1}\), from residual neural dynamics, represented by \(\mathbf{x}^{(2)}_k \in \mathbb{R}^{n_x - n_1}\). We also note that each \(\mathbf{x}_k \) is estimated purely using past neural data (that is, \(y_1,...,y_{t-1}, s_1,...s_{k-1}\)), so \(\mathbf{x}_{k+1} \) has the information about neural modalities up to time \(k\). As such, our model supports causal predictions.

Lines 2-4 of Equation~\ref{eq:equation appendix} describe how the model predicts observations:

\textbf{2}: \(\mathbf{\hat{y}}_t\) is the predicted Gaussian neural signal, combining contributions from both latent states \(\mathbf{x}^{(1)}_k\) and \(\mathbf{x}^{(2)}_k\) through the nonlinear decoders \(\bm{C}_\mathbf{y}^{(1)}(\cdot)\) and \(\bm{C}_\mathbf{y}^{(2)}(\cdot)\). This prediction is performed in Stage 2 and Stage 3 of the architecture (Figure~\ref{architecture}).

\textbf{3}: \(\mathbf{\hat{\lambda}}_k\) represents the predicted firing rate of spike counts, which also combines \(\mathbf{x}^{(1)}_k\) and \(\mathbf{x}^{(2)}_k\) through the nonlinear decoders \(\bm{C}_\mathbf{s}^{(1)}(\cdot)\) and \(\bm{C}_\mathbf{s}^{(2)}(\cdot)\). Notably, unlike the Gaussian decoders \(\bm{C}_\mathbf{y}^{(1)}(\cdot)\) and \(\bm{C}_\mathbf{y}^{(2)}(\cdot)\), our Poisson decoders \(\bm{C}_\mathbf{s}^{(1)}(\cdot)\) and \(\bm{C}_\mathbf{s}^{(2)}(\cdot)\), employ an \(\exp(.)\) nonlinearity output layer activation function to force non-negativity of Poisson rates.
Therefore, Poisson decoders can be expressed as:
\begin{equation}
\begin{aligned}
\bm{C_s}^{(1)}(\cdot) &= \exp \big(\bm{C_s}''^{(1)}(\cdot) \big) \\
\bm{C_s}^{(2)}(\cdot) &= \exp \big(\bm{C_s}''^{(2)}(\cdot) \big) \\
\label{Cs}
\end{aligned}
\end{equation}
where, \(\bm{C_s}''(\cdot)\) has no output layer activation function.
As a result, predicted firing rate can be formulated as:
\begin{equation}
\begin{aligned}
\mathbf{\hat{\lambda}}_k &= \exp \big(\bm{C_s}''^{(1)}(\mathbf{x}_k^{(1)}) + \bm{C_s}''^{(2)}(\mathbf{x}_k^{(2)}) \big)
\label{exp Cs}
\end{aligned}
\end{equation}

This step is also done in Stage 2 and Stage 3 of the architecture (Figure~\ref{architecture}).

\textbf{4}: \(\mathbf{\hat{z}}_k\) is the predicted behavior, decoded from the low-dimensional latents of Stage 1 using the nonlinear decoder \(\bm{C}_\mathbf{z}(\cdot)\) learned during Stage 1 (Figure~\ref{architecture}).

\subsubsection{Learning Stages} \label{app:Learning Parameters}
This section details the structured three-stage optimization procedure used to train \ourmethod{} as described in Equation~\ref{eq:equation appendix}. Each stage is designed to optimize distinct components of the model in order to accurately predict observations while separating behaviorally relevant neural dynamics and residual neural dynamics. Across all stages, the negative log-likelihood (NLL) of the observed data is used as the training objective.

\textbf{Stage 1:}

First, we fit an RNN, composed  of \( {A}^{(1)}(\cdot)\) and modality-specific encoders, \(\mathbf{Enc}^{(1)}_\mathbf{y}(\cdot)\) and \(\mathbf{Enc}^{(1)}_\mathbf{s}(\cdot)\), jointly with the behavior decoder network $\bm{C}_\mathbf{z}$$(\cdot)$, using \(n_1\) states, to minimize behavior prediction NLL given past neural activities \(\mathbf{y}_t\) and \(\mathbf{s}_k\) (Equation~\ref{eq:loss1_stage1_appendix}). This ensures that the neural dynamics that are predictive of behavior are learned. 
The resulting latent states, \(\mathbf{x}_k^{(1)} \), form the first set of latents in \ourmethod{} which are behaviorally relevant latents. The optimization can be expressed as:
\begin{equation}
\left\{
\begin{aligned}
\mathbf{x}_{k+1}^{(1)} &= A^{(1)}( \mathbf{x}_{k}) +\mathbf{Enc}^{(1)}_\mathbf{y}(\mathbf{y}_t) + \mathbf{Enc}^{(1)}_\mathbf{s}(\mathbf{s}_k)\\
\mathbf{\hat{z}}_k &=\bm{C}_\mathbf{z}(\mathbf{x_k^{(1)}}) \\
\mathbf{L}_{z}^{(1)} &= -\sum_{k} \log\big( p_z (\mathbf{z}_k \mid \mathbf{x}_k^{(1)} ) \big)
\label{eq:loss1_stage1_appendix}
\end{aligned}
\right.
\end{equation}
where \(p_z\) represent the conditional likelihood of behavior \(\mathbf{z}_k\) given behaviorally relevant latents in Stage 1.

\textbf{Stage 2:}

Next, we fit two modality-specific decoders to the latents \( \mathbf{x}_k^{(1)} \) to predict the two neural modalities. The objective is to simultaneously minimize NLL of both neural modalities, ensuring accurate predictions across these diverse data types. Since \ourmethod{} accommodates neural modalities from different statistical distributions, the loss function must reflect these differences. To address this, we optimize a weighted combination of the NLLs for the two modalities, incorporating a scaling factor \( \alpha \) to balance the disparity in scales between them (Eq.~\ref{eq:loss2_stage1_appendix}). This \( \alpha \) factor ensures that both modalities contribute meaningfully to the optimization process, regardless of their inherent scale.
\begin{equation}
\left \{
\begin{aligned}
\mathbf{x}_{k+1}^{(1)} &= A^{(1)}( \mathbf{x}_{k}) +\mathbf{Enc}^{(1)}_\mathbf{y}(\mathbf{y}_t) + \mathbf{Enc}^{(1)}_\mathbf{s}(\mathbf{s}_k)\\
\mathbf{\hat{\bar{y}}}_t & = \bm{C}_\mathbf{y}^{(1)}(\mathbf{x}_t^{(1)}) \\
\mathbf{\hat{\bar{\lambda}}}_k & = \bm{C}_\mathbf{s}^{(1)}(\mathbf{x}_k^{(1)}) \\
\mathbf{L}_\mathbf{y,s}^{(1)} & = -\big( \sum_{t} \log \big( p_y (\mathbf{y}_t \mid \mathbf{x}_t^{(1)} ) \big) + \alpha \sum_{k} \log\big( p_s (\mathbf{s}_k \mid \mathbf{x}_k^{(1)} ) \big) \big)
\end{aligned}
\right.
\label{eq:loss2_stage1_appendix}
\end{equation}
Here, $ {p_y}$ and ${p_s}$ represent the conditional likelihoods of the two neural modalities, $\mathbf{y}_t$ and $\mathbf{s}_k$, respectively. This stage ensures that \( \mathbf{x}_k^{(1)} \) adequately explains the behaviorally relevant components of the neural data. Therefore, \(\mathbf{\hat{\bar{y}}}_t\) and \(\mathbf{\hat{\bar{\lambda}}}_k\) correspond to the part of neural modalities \(\mathbf{y}_t\) and \(\mathbf{s}_k\) that are predictable with behaviorally relevant latents.

\textbf{Stage 3:}

In the final stage, we learn the second RNN, composed of \( {A}^{(2)}(\cdot)\) and modality-specific encoders, \(\mathbf{Enc}^{(2)}_\mathbf{y}(\cdot)\) and \(\mathbf{Enc}^{(2)}_\mathbf{s}(\cdot)\) jointly with the modality-specific neural decoders \(\bm{C}_\mathbf{y}^{(2)}(\cdot)\) and \(\bm{C}_\mathbf{s}^{(2)}(\cdot)\) to predict the residual neural dynamics of the two neural modalities that are not already explained by the behaviorally relevant neural latents \(\mathbf{x}^{(1)}_k\). We represent the unpredicted part of multimodal neural activities with \(\mathbf{\tilde y}_t\), \(\mathbf{\tilde s}_k\), such that the full neural predictions are given by:
\begin{equation}
\begin{aligned}
\hat{\mathbf{y}}_t & = \hat{\tilde{\mathbf{y}}}_t +  \hat{\bar{\mathbf{y}}}_t \\
 \hat{\mathbf{s}}_k &\sim \mathcal{P}ois(\mathbf{\hat{\tilde \lambda}}_k \times  \mathbf{\hat{\bar{ \lambda}}}_k)
\label{eq:appendix neural specific}
\end{aligned}
\end{equation}
This final stage ensures that the residual neural dynamics (not explained by \(\mathbf{x}^{(1)}_k\)) are also captured using a separate set of latents. The optimization for this stage is defined as:
\begin{equation}
\left\{
\begin{aligned}
\mathbf{x}_{k+1}^{(2)} &= A^{(2)}( \mathbf{x}_{k}) +\mathbf{Enc}^{(2)}_\mathbf{y}(\mathbf{y}_t) + \mathbf{Enc}^{(2)}_\mathbf{s}(\mathbf{s}_k)\\
\mathbf{\hat{\tilde{y}}}_t &= \bm{C}_\mathbf{y}^{(2)}(\mathbf{x}_t^{(2)}) \\
\mathbf{\hat{\tilde{\lambda}}}_k &= \bm{C}_\mathbf{s}^{(2)}(\mathbf{x}_k^{(2)}) \\
\mathbf{L}_\mathbf{{y,s}}^{(2)} &= -\big(\sum_{t} \log \big( p_y (\mathbf{y}_t \mid \mathbf{x}_t^{(1)}, \mathbf{x}_t^{(2)}) \big) + \alpha \sum_{k} \log\big( {p_s} (\mathbf{s}_k \mid \mathbf{x}_k^{(1)}, \mathbf{x}_k^{(2)}) \big) \big)
\end{aligned}
\right.
\label{eq:loss1_stage2_appendix}
\end{equation}

\subsubsection{Scaling factor \( \alpha \)}\label{alpha_details}
To address the differences in scale between the likelihoods of two neural modalities, we use a scaling factor, 
\( \alpha \), in the loss function to ensure that both neural modalities contribute proportionally to the optimization process.
To determine the value of \( \alpha \), we first compute the average value across all available time steps for each neural modality. This average serves as a baseline, representing the overall scale or magnitude of the observed data for that modality.
Then, assuming the mean value as the predicted value for each modality, we calculate the log-likelihood of the observed data. This step quantifies the "scale" of each modality in terms of their likelihood under their respective distributions.
Finally, the scaling factor \( \alpha \) is determined as the ratio of the likelihoods of the modalities. Specifically, \( \alpha \) is set such that it normalizes the contributions of the modalities, effectively equalizing their influence during the optimization process.

\subsubsection{Hyperparameters and Implementation}\label{hyperparams}
For \ourmethod{}, we employed consistent neural network architectures across both simulated and real-world evaluations. The specific hyperparameters employed for training on the simulation dataset, as well as analyses of the real-world dataset, are detailed in Table~\ref{tab:hyperparams_sim}. For encoders and decoders in our model i.e., \(\mathbf{Enc}^{(1)}_\mathbf{y}(\cdot)\), \(\mathbf{Enc}^{(1)}_\mathbf{s}(\cdot)\),  \(\mathbf{Enc}^{(2)}_\mathbf{y}(\cdot)\),  \(\mathbf{Enc}^{(2)}_\mathbf{s}(\cdot)\), \(\bm{C}^{(1)}_\mathbf{y}(\cdot)\), \(\bm{C}^{(1)}_\mathbf{s}(\cdot)\), \(\bm{C}^{(2)}_\mathbf{y}(\cdot)\), \(\bm{C}^{(2)}_\mathbf{s}(\cdot)\) we used multi-layer perceptrons (MLPs) with each having specific number of hidden layers and neurons (Table~\ref{tab:hyperparams_sim}). 
Otherwise, for our \lourmethod{} analysis, we remove the hidden layers and nonlinearity in feedforward neural networks. Notably, unlike the Gaussian decoders \(\bm{C}_\mathbf{y}^{(1)}(\cdot)\) and \(\bm{C}_\mathbf{y}^{(2)}(\cdot)\), our Poisson decoders \(\bm{C}_\mathbf{s}^{(1)}(\cdot)\) and \(\bm{C}_\mathbf{s}^{(2)}(\cdot)\) employ an \(\exp(.)\) nonlinearity output layer activation function to force non-negativity of Poisson rates.
We adopted a Step Decay learning rate scheduler~\cite{ge2019step}, which is known for its effectiveness in training deep neural networks. It was initialized with a learning rate of \(10^{-3}\), which was halved (\( \gamma = 0.5\)) every 400 steps. To ensure stable training and convergence, we initialized all model parameters using the Xavier-normal initialization method \cite{glorot2010understanding}. For optimization, we utilized the AdamW optimizer \cite{loshchilov2017decoupled}. We trained the model for a maximum of 1000 epochs to ensure thorough learning, while employing early stopping to avoid overfitting. All models were trained on CPU servers equipped with AMD EPYC 7513 and 7542 processors (2.90 GHz, 32 cores) with parallelization.

In simulation experiments, the state dimensions were aligned with those of the true underlying model. For real-world data analyses, we explored the impact of varying the state dimensions \(n_1\) and \(n_x\) in one session of data. Specifically, we tested dimensions \(n_1, n_x \in [2, 4, 8, 16, 32, 64, 128]\) and selected \(n_1=16\) and \(n_x=64\) based on behavior decoding and neural prediction accuracy on a held-out session. This choice is a model-selection criterion rather than an estimate of the intrinsic dimensionality of behaviorally relevant neural activity. 
\vskip 0.15in
\begin{table}[!h]
  \caption{\textbf{Hyperparameter settings of \ourmethod{}.} Encoders and decoders architectures, as well as training configurations, used across all experiments and models.}
  \label{tab:hyperparams_sim}
  \vskip 0.10in
  \centering
  \begin{tabular}{l c}
    \toprule
    \textbf{Hyperparameter} & \textbf{Value} \\
    \midrule
    Encoders (\(\mathbf{Enc_y}, \mathbf{Enc_s}\))  & [64, 128] \\
    Neural decoders (\(\bm{C}_\mathbf{y}, \bm{C}_\mathbf{s}\))            & [128, 128, 128] \\
    Behavior decoder (\(\bm{C}_z\))               & [64] \\
    Initial learning rate                               & \(10^{-3}\) \\
    LR scheduler                                        & StepLR \\
    Batch size                                          & 32 \\
    Number of epochs                                    & 1000 \\
    Weight decay                                        & \(10^{-3}\) \\
    \bottomrule
  \end{tabular}
  \vskip 0.10in
\end{table}
\vskip 0.15in
\subsubsection{Evaluation Metrics}\label{app:evaluation metric}

After learning \ourmethod{}, we utilize the learned nonlinear modality-specific encoders \( \mathbf{Enc}^{(1)}_\mathbf{y}(\cdot) \), \(\mathbf{Enc}^{(1)}_\mathbf{s}(\cdot)\), \( \mathbf{Enc}^{(2)}_\mathbf{y}(\cdot) \), \(\mathbf{Enc}^{(2)}_\mathbf{s}(\cdot)\), and recursion networks \( A^{(1)}(\cdot) \), \( A^{(2)}(\cdot) \) to infer 1-step-ahead predicted latent states using Equation~\ref{eq:equation_formulation_expanded} for the held-out test data in cross-validation. Using these inferred latent states, we compute the predicted neural activity and behavior by applying the corresponding decoders \( \bm{C}_\mathbf{y}(\cdot) \), \(\bm{C}_\mathbf{s}(\cdot)\), and \( \bm{C}_\mathbf{z}(\cdot) \). For behavior predictions, we refer to this process as \emph{decoding}, as the model exclusively utilizes the neural modalities \(\mathbf{y}_t\) and \(\mathbf{s}_k\), and does not incorporate behavior itself as an input during inference. 

To evaluate model performance, we use \( 5 \)-fold cross-validation for both simulation and real data analyses. For LFP neural activity and continuous behavioral signals, which are modeled using a Gaussian distribution, we report the Pearson Correlation Coefficient (CC) in all of our analysis (Figs.~\ref{fig:sabes_results_all},~\ref{fig:nhp basleines barplot},~\ref{fig:ablation_figure}) between predicted and actual observations. 

For spike count data, modeled with a Poisson distribution, we quantify one-step-ahead prediction accuracy using the Prediction Power (PP) metric, defined as \( \mathbf{PP = 2 \, AUC - 1} \), where \( \mathbf{AUC}\) is the area under the curve of the receiver operating characteristic (ROC) curve~\cite{macke2011empirical}. The ROC curve is constructed by using the one-step-ahead predicted firing rates (\( \mathbf{\hat{\lambda}}_{k+1} \)) as classification scores to determine whether a time step contains a spike~\cite{truccolo2010collective}. 

\subsection{Baselines and model variants} \label{baselines}
We compare \ourmethod{} against multimodal baselines and against ablated variants of our own model. All models use the same cross-validation protocol and the same number of channels per modality, and we performed hyperparameter sweeps (including latent dimensionality) for each baseline while keeping the total number of trainable parameters comparable across methods, so that gains are not confounded by model capacity.

\textbf{Model variants.} \lourmethod{} keeps the three-stage architecture but replaces every encoder and decoder with linear maps. \uourmethod{} runs Stage 3 only, learning neural dynamics without behavior supervision (equivalently $n_1=0$) and fitting a behavior decoder downstream on the resulting latents. \ourmethod{} w/o Stage 3 keeps Stages 1--2 only and therefore has no residual neural latents. \ourmethod{} Joint trains all latents in a single stage with a combined behavior-decoding and neural-prediction loss. Single-modal variants remove the encoder and decoder of one neural modality, so no fusion occurs during inference.

\textbf{External baselines.} MSID~\citep{ahmadipour2024multimodal} is a linear multiscale subspace-identification method; we used the authors' implementation with horizons $h_y=h_z=10$ as specified in the original work. MVAE~\citep{wu2018multimodal} is a generic multimodal variational autoencoder without temporal modeling or behavior supervision; we treated each time step as an independent sample and fit linear regressions on the inferred latents, and it cannot perform time-series prediction. CEBRA~\citep{schneider2023learnable} is a convolutional encoder trained with a contrastive objective; we used its multi-session mode to process LFP and spikes jointly and fit linear readouts on the embeddings, since CEBRA has no decoder. mmPLRNN~\citep{kramer2021reconstructing} uses piecewise-linear RNN dynamics with variational inference; we added a Poisson observation model following the authors' Appendix C. MMGPVAE~\citep{gondur2023multi} places Gaussian-process priors over the latents and optimizes the ELBO; we used 64-dimensional latents per modality with 32 shared and matched its encoder/decoder architecture to ours. For all external baselines we used the authors' public implementations and recommended hyperparameters.

Neither mmPLRNN nor MMGPVAE supports predicted (one-step-ahead) inference, so for a fair comparison we modified their inference to use only neural observations up to time $k-1$ when inferring the latent state at $k$, then decoded the predicted observations from those latents. Both were additionally trained in a supervised variant in which a feedforward map from the inferred latents to behavior, and its reconstruction loss, were added to the original objective; behavior is never an input at inference. Unlike \ourmethod{}, this makes their latent space more behaviorally informative overall but does not separate behaviorally relevant from residual neural dynamics.

\subsection{NHP grid reaching dataset}\label{sabes_data}
We analyzed a publicly available dataset~\cite{o_doherty_2020_3854034} in which a macaque monkey (Monkey I) performed a grid reaching task. Neural activity was recorded from the primary motor cortex (M1) using a 96-channel electrode array as the subject controlled a 2D cursor to reach randomly appearing targets on a grid within a virtual reality environment. Targets were presented sequentially, with no intervening time gaps between their appearances. In our analysis, we took the subject’s 2D cursor velocity recorded at a 10 ms timescale as the behavior time-series to decode. 

For the spiking data, we utilized multi-unit spiking activity recorded at a 10 ms timescale. From the four recording sessions, we selected the top 20 spiking channels based on their behavior prediction accuracies. For the local field potential (LFP) data, signals were extracted from the raw neural recordings using a low-pass filter with a 300 Hz cut-off frequency. The filtered signals were downsampled to 100 Hz (10 ms timescale). Similar to the spiking data, we identified the top 20 LFP channels across the four sessions based on their behavior prediction accuracies. We report the mean and SEM computed across four sessions and five cross-validated folds (N=20).
\subsection{Simulation Details} \label{app:sim_apdx}
For our simulation data in section~\ref{app:simulation_section}, we generated Gaussian and Poisson observations from four randomly initialized systems as defined by the dynamics in 

\begin{equation}
\begin{aligned}
\mathbf{x}_{k+1} &= F(\mathbf{x}_{k}) + \mathbf{w}_k \\
\mathbf{y}_k \mid \mathbf{x}_k &\sim \mathcal{N} \big (\mathbf{y}_k\mid \bm{C}_\mathbf{y}(\mathbf{x}_{k})\big) \\
\mathbf{s}_k \mid \mathbf{x}_k &\sim \mathcal{P}ois \big(\mathbf{s}_k\mid\bm{C}_\mathbf{s}(\mathbf{x}_{k})\big) \\
\mathbf{z}_k \mid \mathbf{x}_k &\sim \mathcal{N}\big(\mathbf{z}_k\mid\bm{C}_\mathbf{z}(\mathbf{x}_{k})\big) \\
\end{aligned}
\label{eq:dynamics_simulation}
\end{equation}

In this setup, the latent state evolves in time according to a recursion function \(F(\cdot)\) with additive noise \(\mathbf{w}_k\). \( \mathbf{y}_k \) represents Gaussian neural activity, \( \mathbf{s}_k \) corresponds to spiking neural activity, and \( \mathbf{z}_k \) denotes behavior. A total of eight latent variables (\( n_x = 8 \)) were drawn from the system, where \( n_1 = 4 \) latent dimensions were allocated to the behaviorally relevant subspace and the remaining \( n_2 = n_x - n_1 = 4 \) to the behaviorally independent subspace. These latent variables were then mapped to their corresponding observations using nonlinear functions \(\bm{C}_\mathbf{y}(\cdot)\), \(\bm{C}_\mathbf{s}(\cdot)\), and \(\bm{C}_\mathbf{z}(\cdot)\). For each randomly initialized system, we generated a total of \(2\times10^{6}\) samples. 
These simulation settings demonstrate the benefits of fusing multimodal information from two neural modalities while separating the behaviorally relevant neural latents and residual neural latents, leading to improved predictions of both neural and behavior time-series. 
\clearpage
\subsection{Supplementary Results}

\subsubsection{Our method fuses multimodal data, thus better predicting all modalities in simulations} \label{app:simulation_section}
We first validated the ability of \ourmethod{} to effectively fuse information across multiple modalities through a series of simulation experiments. These simulations involved three modalities with shared and independent dynamics. The underlying dynamical system used to generate the synthetic time-series for these modalities is defined in Equation~\ref{eq:dynamics_simulation}. 
We set the total simulated neural latent dimensionality to \(n_x = 8\), with four dimensions allocated to behaviorally relevant latents (\(n_1 = 4\)) and the remaining four to residual neural latents (\(n_2 = 4\)).

\begin{figure}[ht]
\begin{center}
\centerline{\includegraphics[width=\textwidth]{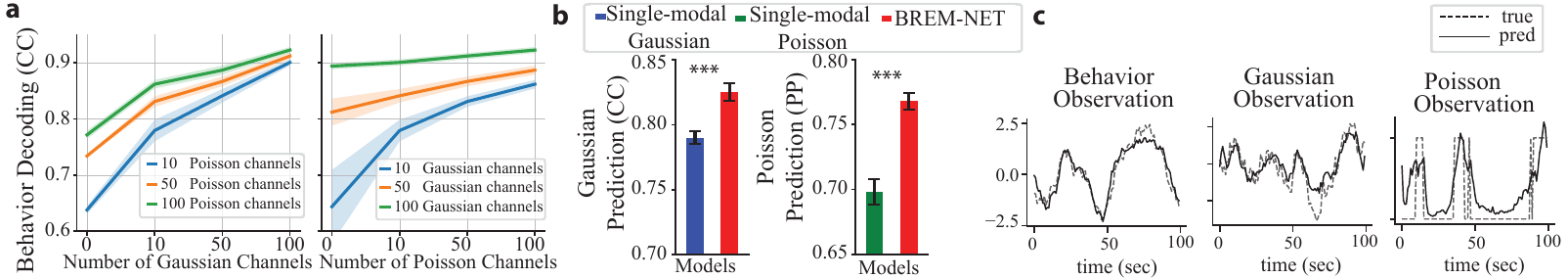}}
\caption{\textbf{\ourmethod{} improves behavior decoding and multimodal neural predictions in a simulated dataset.} Cross-validated accuracy of behavior decoding and Gaussian and Poisson neural prediction in four simulated systems (N=20). \textbf{(a)} \emph{left}: behavior decoding correlation coefficient (CC) when 10, 50, or 100 Poisson channels were the primary modality and an increasing number of Gaussian channels were fused with them. Lines represent the mean and shaded areas show the standard error of the mean (SEM). \emph{right}: Similar to \emph{left} when Gaussian channels were the primary modality. \textbf{(b)} \emph{left}: Prediction CC of  Gaussian modality using \ourmethod{} vs. when training a single-modal model with Gaussian modality alone. Bars represent the mean and error bars show the SEM. \emph{right}: Prediction Power (PP) of the Poisson modality using \ourmethod{} vs. when training a single-modal model on the Poisson modality alone.  Asterisks indicate significance of comparison (***: $p < 10^{-7}$ one-sided Wilcoxon signed-rank test). \textbf{(c)} True versus predicted time-series using
\ourmethod{}, showing the model’s ability to accurately predict all observation modalities.}
\label{fig:simulation_results}
\end{center}
\end{figure}

We generated observations from both Gaussian and Poisson distributions using 100 channels each and compared \ourmethod{} to single-modal models that only used either Gaussian or Poisson observations alongside behavior (Appendix~\ref{baselines}).
Specifically, in Figure~\ref{fig:simulation_results}a, we chose a primary neural modality, Poisson in the left panel and Gaussian on the right panel. We then fused this primary modality with gradually increasing numbers of channels of the other secondary modality, as shown on the x axis. As such, the points corresponding to 0 on the x axis indicate the single-modal performance and all other points show \ourmethod{} that fuses different numbers of channels of the two different modalities.

We found that \ourmethod{} demonstrated superior performance in behavior decoding compared to single-modal models, as shown in Figure~\ref{fig:simulation_results}a. In scenarios where the Poisson modality served as the primary modality, incorporating Gaussian channels consistently improved behavior decoding accuracy (Figure~\ref{fig:simulation_results}a, left). Similar trend was observed when the Gaussian modality was the primary modality and Poisson channels were fused (Figure~\ref{fig:simulation_results}a, right). Further, incorporating information from one neural modality enhanced the prediction accuracy of the other modality, showing the power of cross-modality information fusion (Figure~\ref{fig:simulation_results}b). 
Consistently, \ourmethod{} successfully predicted both neural time-series modalities and behavioral time-series from the learned latent states (Figure~\ref{fig:simulation_results}c). 

\vskip 0.2in

\subsubsection{Robustness to missing data, temporal resolution mismatch, and dropout}\label{different resolution}
In practical neural recording scenarios, different modalities—such as LFPs and spikes—are often sampled at different temporal resolutions due to hardware limitations or preprocessing constraints. Additionally, missing data are common in real-world datasets due to noise, sensor dropout, or transmission errors. Although the dataset used in our experiments is fully observed, \ourmethod{} is explicitly designed to handle missing data and asynchronous sampling as explained in Methods section~\ref{methods section}. This is achieved through modality-specific encoder networks, which flexibly incorporate available modalities at each time step. If a modality is missing, its encoder output can be set to zero, allowing the latent dynamics to update using only the available inputs.

To empirically test robustness to mismatched sampling rates, we downsampled the LFP signals by a factor of 5 in the NHP grid-reaching dataset, while leaving the spike and behavior data unchanged. This setup mimics real-world conditions where continuous signals are collected at lower frequencies than discrete spike trains, or where LFP observations may be intermittently unavailable. As shown in Table~\ref{tab:diff_timescales}, behavior decoding and spike prediction performance remained largely stable, whereas LFP prediction degraded due to masked predictions on downsampled timesteps. This confirms that \ourmethod{} is resilient to both temporal resolution mismatches and partial observability, effectively integrating modalities sampled at different rates without requiring strict alignment.

In addition to resolution mismatches, we also evaluated robustness under irregular and stochastic missing data. At each timestep, either spikes or LFPs were randomly dropped with probability \(q \in \{0.2, 0.4, 0.6, 0.8\}\), creating asynchronous gaps where only one modality was available. Unlike the resolution mismatch experiment, where missing samples occur regularly, this dropout analysis simulates challenges such as channel noise, transient artifacts, or partial recording failures that occur irregularly. As shown in Table~\ref{tab:dropout_results}, decoding performance remained remarkably stable across dropout probabilities, particularly for LFPs. Even when 80\% of LFP samples were removed, behavior decoding declined by less than 2\%. Spike dropouts produced a larger impact, consistent with spikes carrying richer behavior-related information in this dataset (Fig.~\ref{fig:sabes_results_all}c). Still, the performance degradation remained modest, with about a 5\% drop at 40\% spike dropout. Together, these analyses demonstrate that \ourmethod{} flexibly integrates multimodal inputs under asynchronous sampling and partial observability, maintaining robust decoding accuracy under realistic and adverse recording conditions.

\begin{table*}[!ht]
    \vskip 0.10in
  \caption{\textbf{\ourmethod{} is robust to temporal resolution mismatch and missing observations.} We compare \ourmethod{} trained under two conditions: (1) \textbf{same-resolution}, where both LFP and spike data are aligned and sampled at the same rate, and (2) \textbf{different-resolution}, where LFP signals are downsampled by a factor of 5 to simulate asynchronous sampling. These results demonstrate that \ourmethod{} can effectively integrate multimodal signals under realistic constraints such as sampling mismatch and missing observations without compromising decoding performance.}
  \label{tab:diff_timescales}
    \vskip 0.10in
  \centering
  \begin{tabular}{lcccc}
    \toprule
    \textbf{Models} & \textbf{Behavior Decoding} & \multicolumn{2}{c}{\textbf{Neural prediction}} \\
    \cmidrule(lr){3-4}
    & (CC) & \textbf{LFP (CC)} & \textbf{Spike (PP)}\\
    \midrule
    \makecell[l]{\ourmethod{} \\ w/ different-resolution} & 0.7571 $\pm$ 0.0051 & 0.6272 \(\pm\) 0.0234 & 0.3701 \(\pm\) 0.0031 \\
    \makecell[l]{\ourmethod{} \\ w/ same-resolution} & 0.7645 $\pm$ 0.0052 & 0.8078 $\pm$ 0.0125 & 0.3746 $\pm$ 0.0036 \\
    \bottomrule
  \end{tabular}
    \vskip 0.10in
\end{table*}
\vskip 0.10in
\begin{table}[th]
\centering
\caption{\textbf{\ourmethod{} is robust to dropouts.} 
We evaluated decoding performance when either spikes or LFPs were randomly dropped at each timestep with probability $q \in \{0.2, 0.4, 0.6, 0.8\}$.
\ourmethod{} remained highly resilient, particularly to LFP dropouts. Spike dropouts produced a larger effect, consistent with their primary role in behavior decoding.
These results demonstrate that \ourmethod{} flexibly integrates asynchronous and incomplete modalities while maintaining robust behavior decoding performance.}
\label{tab:dropout_results}
   \vskip 0.10in
\begin{tabular}{ccc}
\toprule
\textbf{Dropout probability} & \multicolumn{2}{c}{\textbf{Behavior Decoding (CC)}} \\
\cmidrule(lr){2-3}
 & \textbf{LFP dropout} & \textbf{Spike dropout} \\
\midrule
0.0 & 0.7645 $\pm$ 0.0052 & 0.7645 $\pm$ 0.0052 \\
0.2 & 0.7613 $\pm$ 0.0061 & 0.7543 $\pm$ 0.0042 \\
0.4 & 0.7601 $\pm$ 0.0071 & 0.7245 $\pm$ 0.0063 \\
0.6 & 0.7573 $\pm$ 0.0067 & 0.6993 $\pm$ 0.0072 \\
0.8 & 0.7516 $\pm$ 0.007 & 0.6292 $\pm$ 0.0065 \\
\bottomrule
\end{tabular}
\end{table}
\subsubsection{Functional Separation of behaviorally relevant and residual neural latents}\label{app:separation}

\begin{figure}[ht]
\begin{center}
\centerline{\includegraphics[width=\textwidth]{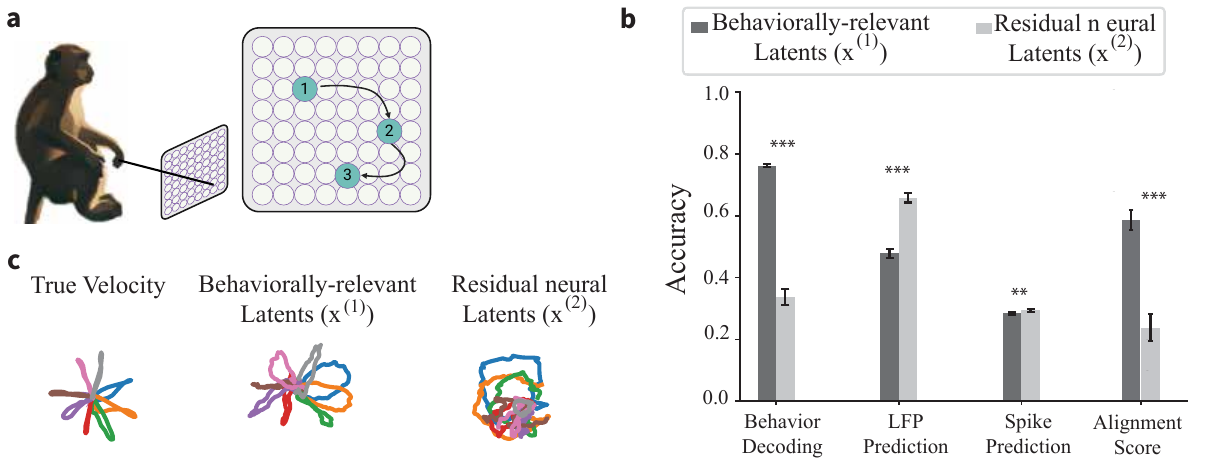}}
\caption{\textbf{Separation of behaviorally relevant and residual neural dynamics through structured multi-stage learning.} Evaluation of the separation of the learned latent states by using $\mathbf{x}^{(1)}$ and $\mathbf{x}^{(2)}$ separately to predict behavior and neural modalities.
\textbf{(a)} NHP grid-reaching dataset. \textbf{(b)} Behavior decoding and alignment score are substantially higher when using the behaviorally relevant latents $\mathbf{x}^{(1)}$ compared to the residual neural latents $\mathbf{x}^{(2)}$ \if (0.7629 vs. 0.3367)\fi. 
Conversely, neural prediction accuracy improves when using $\mathbf{x}^{(2)}$, demonstrating its specialization in capturing residual neural variability. Together, these results indicate a functional specialization: $\mathbf{x}^{(1)}$ carries the behaviorally relevant dynamics learned in Stage 1, while $\mathbf{x}^{(2)}$ captures the residual neural dynamics learned in Stage 3 that improve neural prediction.
 \textbf{(c)} True velocity trajectories are much more aligned with behaviorally relevant neural
latent trajectories than with residual neural latent trajectories.
\if 0\textbf{(c–d)} same as a-b for NHP center-out reaching dataset with \(n_1=4\), \(n_x=16\). Asterisks indicate significance of comparison (***: $p < 10^{-7}$, **: $p < 0.001$ one-sided Wilcoxon signed-rank test)\fi}
\label{app:x1_x2_both_preds}
\end{center}
\end{figure}

\clearpage


\subsubsection{Ablation studies}\label{app:ablations}
We ablated the components of \ourmethod{} on the NHP grid reaching dataset (Fig.~\ref{fig:ablation_figure}; see Appendix~\ref{baselines} for variant definitions). Multimodal fusion, behavior supervision and the staged learning procedure each contribute: the behaviorally relevant latents of the full model are more aligned with behavior than those of the single-modal variants, of \uourmethod{}, and of the joint-loss variant, and \ourmethod{} matches or exceeds every variant on behavior decoding and neural prediction. Removing Stage 3 leaves behavior decoding intact but substantially degrades neural prediction, confirming that the residual neural latents carry residual neural variability.

\begin{figure}[!ht]
\begin{center}
\centerline{\includegraphics[width=\textwidth]{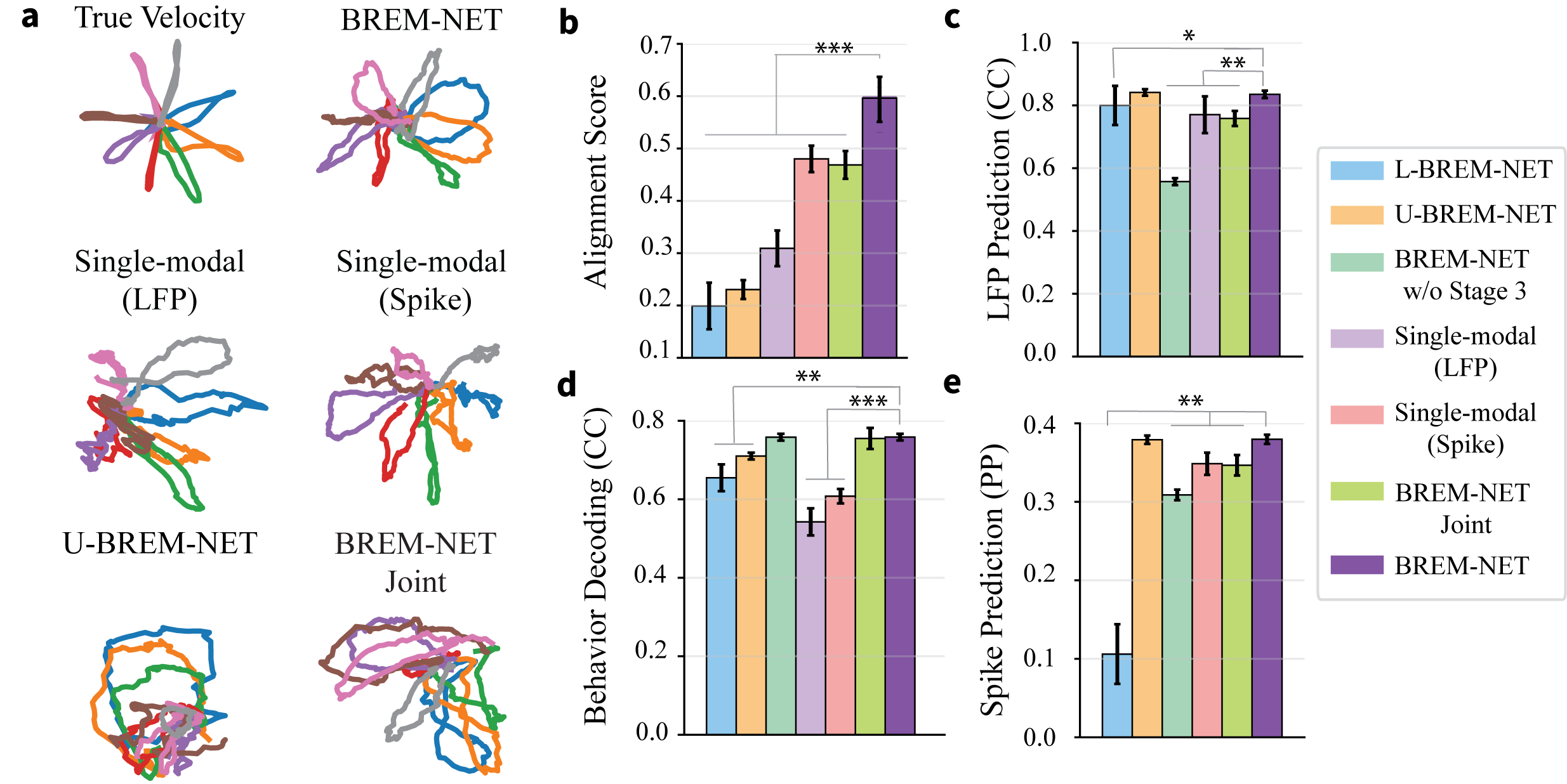}}
\caption{\textbf{Ablations show that multimodal structured modeling improves the accuracy and behavior alignment of behaviorally relevant neural latents while also capturing residual neural variability.} \textbf{(a)} Condition-averaged behaviorally relevant latent trajectories. \textbf{(b)} Alignment score. \textbf{(c)} LFP prediction CC. \textbf{(d)} Behavior decoding CC. \textbf{(e)} Spike prediction PP. Bars show mean $\pm$ SEM across all sessions and 5 folds (N=20); asterisks indicate significance.}
\label{fig:ablation_figure}
\end{center}
\end{figure}

\end{document}